\documentclass[final,3p]{elsarticle}

\usepackage{lineno,hyperref}
\usepackage{graphicx}
\usepackage{caption}
\usepackage{multirow}
\usepackage{color}
\usepackage[table]{xcolor}
\usepackage{setspace}
\modulolinenumbers[5]

\journal{Journal of \LaTeX\ Templates} 

\begin{document}

\begin{frontmatter}

\title{Extractive Multi-document Summarization Using Multilayer Networks}

\author{Jorge V. Tohalino, Diego R. Amancio}
\address{Institute of Mathematics and Computer Science, University of S\~ao Paulo \\ S\~ao Carlos, S\~ao Paulo, 13566-590,
Brazil}




\onehalfspacing

\begin{abstract}
Huge volumes of textual information has been produced every single day. In order to organize and understand such large datasets, in recent years, summarization techniques have become popular. These techniques aims at finding relevant, concise and non-redundant content from  such a big data. While network methods have been adopted to model texts in some scenarios, a systematic evaluation of multilayer network models in the multi-document summarization task has been limited to a few studies.
Here, we evaluate the performance of a multilayer-based method to select the most relevant sentences in the context of an extractive multi document summarization (MDS) task.  In the adopted model, nodes represent sentences and  edges are created based on the number of shared words between sentences. Differently from previous studies in multi-document summarization, we make a distinction between edges linking sentences from different documents (inter-layer) and those connecting sentences from the same document (intra-layer).  As a proof of principle, our results reveal that such a discrimination between intra- and inter-layer in a multilayered representation is able to improve the quality of the generated summaries. This piece of information could be used to improve current statistical methods and related textual models.
\end{abstract}


\end{frontmatter}


\doublespacing

\section{Introduction}

Since textual information available on the Internet has increased in recent years, methods that classify, understand and present the information in a clear and concise way have played a prominent role in data and text mining applications~\cite{Manning:1999:FSN:311445,0295-5075-109-3-30005}. Automatic summarization, question answering and information retrieval~\cite{Manning:1999:FSN:311445} are some of the multi-way proposed solutions to address the problem of managing large volumes of unstructured data, such as written texts.

Automatic summarization is the process of creating a compressed version (summary) of one (single-document summarization) or more documents (multi-document summarization (MDS)) by extracting the most important content~\cite{ferreira}. Automatic summarization techniques are traditionally divided into two groups: extractive and abstractive summarization. The task of extractive summarization consists in concatenating several sentences which are selected without modification, i.e. exactly as they appear in the original document. On the other hand, the creation of abstractive summaries is a more complex and difficult task, because it involves paraphrasing sections of the source document and, for this reason, it requires natural language generation tools. In addition, abstractive methods may reuse clauses  or phrases from original documents~\cite{Nenkova:2011}.  In this work, we target our analysis on extractive summarization applied to a set of documents (MDS).  

The most traditional employed methods to select relevant sentences for extractive summarization are divided into the following major classes: methods based on word frequency, sentence clustering and machine learning. In recent years, a new class of methods based on network theory have been proposed to analyze texts.
Applications of network models in text analysis include the study of scientific documents~\cite{10.1371/journal.pone.0187164,SILVA2016487,VIANA2013371,Chen:2005:CPP:1040830.1040859}, stylometry~\cite{AMANCIO20124406,MEHRI20122429,1742-5468-2015-3-P03005,10.1371/journal.pone.0136076}, sense discrimination~\cite{Agirre:2009:PPW:1609067.1609070,0295-5075-98-5-58001} and several other applications~\cite{YU20111370,LIU20083048}. The problem of creating single-document extractive summaries has benefited from these previous network models of texts~\cite{lucas,Ribaldo2012,diego}. It has been claimed that network features overcome other traditional statistical methods when they are used to identify the most central sentences~\cite{lucas,Ribaldo2012}. While most of the studies applying network concepts in summarization have been limited to the single-document counterpart, here we evaluate the usefulness of networks in the multi-document scenario. In particular, the main objective of this paper is to probe whether a discrimination between intra- and inter-layer edges is able to improve the characterization of documents modeled as multilayer complex networks.  This is an important feature to be considered in the models because a sentence connected to many other sentences from other documents may indicate a high relevance of the approached topics.



In the adopted method, nodes represent sentences and edges are established according to the lexical similarity between two sentences. A multilayer network is created by considering each document as a layer. As such, two types of links arises: those connecting sentences from the same documents and the links connecting sentence from distinct sources.  In addition to the traditional network measurements, we also used dynamical measurements to improve the characterization of the obtained networks. An evaluation on three corpora (English and Portuguese) revealed that a simple distinction between intra- and inter-layer edges yields better performance in comparison to network methods not relying upon multilayered representations.
A complimentary analysis also revealed that, in general, traditional measurements such as degree and PageRank  yields a good performance for the MDS task. Finally, we also found that, differently from the single-document case~\cite{lucas}, there is no strong correlation among the evaluated network measurements, which suggests that they could be combined to improve the identification of relevant sentences.

This paper is organized as follows: In Section \ref{section:related}, we present a brief survey of works that used complex networks for extractive summarization.  The description of the adopted network model, and the network metrics we used to rank sentences are described in Section \ref{section:metodologia}. In Section \ref{section:results}, the results are presented and discussed. Finally, the conclusions and prospects for future work are discussed in Section \ref{section:conclusions}.

\section{Related Work}\label{section:related}

Several works have addressed the task of extractive summarization based on complex networks tools and methods. For example, in the work of Ribaldo et al.~\cite{Ribaldo2012}, the MDS task  was applied for documents in Brazilian Portuguese. The authors first extracted all sentences from the cluster of documents and then they modelled them as a single network. After the pre-processing stage, sentences were represented as nodes, which were linked by traditional similarity measurements. In order to select the best ranked sentences, the authors used simple network measurements, including degree, clustering coefficient and average shortest path length. A simple heuristic to avoid redundant sentences in the generated summaries was also applied. The results showed that the proposed method yielded competitive results, which were close to the best statistical systems available for the Portuguese language.  Even though this work addressed the MDS with a graph-based approach, no distinction between intra- and inter-layer edges was considered.

Antiqueira et al.~\cite{lucas} represented sentences as nodes, which are linked   if they share significant lemmatized nouns. The authors applied static complex network metrics to identify the relevant sentences to compose the extract. A summarization system based on voting system was used to combine the results of summaries generated by different measurements. Some systems achieved good results, which are comparable to the top single-document summarizers for Brazilian Portuguese.

Leite and Rino~\cite{Leite} used multiple features to automatically select the best attributes from single-layer complex networks and other linguistic features. More specifically, the authors combined $11$ linguistic features and $26$ features network-based measurements. For extract generation, Leite and Rino used machine learning to classify each sentence as present or not present in summary. An evaluation in a corpus o Portuguese texts confirmed that the proposed network methods can be combined with linguistic features so as to improve the characterization of textual documents.

In the work of Erkan and Radev~\cite{Erkan:2004}, sentences are represented as nodes.  The bag-of-words model is used to represent the sentences. A connection between two nodes is established if the cosine similarity between the vectors of sentences is higher than a predefined threshold. To rank the sentences, the authors used degree centrality and eigenvector based metrics. Competitive results were reported, even if when applied to noisy data.

In the work of Mihalcea~\cite{Mihalcea:2005}, similarly to other studies, sentences are nodes and edges represent the lexical similarity between sentences. The authors used recommendation algorithms for Web Pages to select the most informative sentences. The proposed algorithms used both Google's PageRank \cite{Pageetal98} and HITS~\cite{Kleinberg:1999}. Mihalcea~\cite{Mihalcea:2005} considered three network types: undirected, forward (edges reflecting the natural reading flow) and backward (edges going from the current to the previous word). The systems were evaluated by using the English corpus DUC-2002~\cite{Over2002}. The best performance was achieved with the HITS algorithm for English texts. Conversely, for the Portuguese scenario, the PageRank algorithm yielded the best performance.

\section{Methodology} \label{section:metodologia}

In the current paper, we propose a method based on complex network measurements for multi-document summarization  by modeling a set of documents as a multilayer network. We represent each document sentence by a node and edges are established based on the cosine similarity between two sentences.  The adopted method to extract the most relevant sentences from the texts can be divided into the following steps: document pre-processing, sentence vectorization, network creation, measurements extraction and summarization (i.e. sentence selection). These steps are detailed in sections~\ref{pre-pro}--\ref{sum-sel}.

\subsection{Datasets}

The datasets used in this work comprises texts originally written in Portuguese and English. Since in a multi-document context texts are organized according to the subject approached, each dataset is organized in a set of clusters of related texts. The details of the datasets are provided below:

\begin{itemize}

\item CSTNews for Portuguese MDS~\cite{Ribaldo2012}: this corpus includes documents extracted from on-line Brazilian news agencies: \emph{Folha de S\~ao Paulo}, \emph{Estad\~ao}, \emph{O Globo}, \emph{Gazeta de Povo} and \emph{Jornal do Brazil}. This corpus comprises 140 news items, which are divided into 50 clusters. Each cluster contains 2 or 3 documents sharing the same topic. This corpus includes two reference manual multi-document summaries for each cluster. Each summary has a 70\% compression rate.

\item DUC-2002 for English MDS~\cite{Over2002}: this corpus comprises a set of 567 texts divided into 59 clusters from the following on-line news journals: \emph{Wall Street Journal}, \emph{AP Newswire}, \emph{San Jose Mercury News}, \emph{Financial Times}, \emph{LA Times} and \emph{FBIS}. Each cluster include two 200-word reference summaries.

\item DUC-2004 for English MDS~\cite{duc2004}: this corpus contains 50 clusters of 10 documents each. Four human reference summaries with 665 characters length were produced for this corpus. The DUC-2004 documents were extracted from the following sources: \emph{Associated Press Newswire}, \emph{New York Times Newswire}, and \emph{Xinhua News Agency}.

\end{itemize}

\subsection{Document pre-processing} \label{pre-pro}

In this step, the following pre-processing steps are applied:  text segmentation, removal of unnecessary words and lemmatization.
In text segmentation, sentences boundaries are recognized. We defined a sentence as any text segment separated by a period, exclamation or question mark. 
Punctuation marks and stopwords (such as articles and prepositions) were also removed. Finally, we lemmatized the remaing words so as to map the remaining words to their canonical forms. As a consequence, plural nouns and conjugated verbs are transformed to their singular and infinitive forms. To illustrate this process, we provide in Table \ref{tab:procesamiento} a small piece of text undergoing the aforementioned pre-processing steps.

\begin{table*}[ht]
\centering
\caption{Example of pre-processing steps applied to a piece of text extracted from Wikipedia. We first show the original document divided into seven sentences (a--g). The corresponding preprocessed sentences are shown in lines 1 -- 7.
}
\label{tab:procesamiento}
\begin{tabular}{|l|}
\hline
\multicolumn{1}{|l|}{\textbf{Original Sentences}} \\ \hline
a. Arequipa is the capital and largest city of the Arequipa Region from Peru \\ 
b. It is Peru's second most populous city with 861,145 habitants \\ 
c. Arequipa is the second most industrialized and commercialized city in Peru \\ 
d. Its industrial activity includes manufactured goods and camelid wool products for export \\ 
e. The city has close trade ties with Chile, Bolivia and Brazil \\ 
f. The city was founded on August 15, 1540, by Garcí Manuel de Carbajal  \\ 
g. The historic center of Arequipa spans an area of 332 hectares and is a UNESCO World Heritage Site \\ \hline
\multicolumn{1}{|l|}{\textbf{Pre-proccesed Sentences}}  \\ \hline
1. arequipa  capital large city arequipa region peru  \\ 
2. peru second populous city habitant  \\ 
3. arequipa second industry commerce city peru  \\ 
4. industry activity include manufacture good camelid wool product export  \\ 
5. city trade tie chile bolivia brasil  \\ 
6. city be found august garci manuel carbajal \\ 
7. history center arequipa span area unesco world heritage site \\ \hline
\end{tabular}
\end{table*}

\subsection{Sentence vectorization}

The next step in mapping a text into a network is the sentence vectorization.
The tf-idf (term frequency-inverse document frequency) weighting is a widely used method for document representation. We used this method because it was employed with satisfactory results in several other NLP tasks~\cite{robertson2004}. The term frequency (tf) of a term $t$ in a document $d$ is calculated as $\textrm{tf}(t,d) = {f(t,d)}/{|d|},$
where $f(t,d)$ is the number of times the term $t$ appears in the document $d$ and $|d|$ is the total number of terms in $d$. The inverse document frequency (idf) of term $t$ in the collection of documents $D$ is calculated as
\begin{equation}
	\textrm{idf}(t,D) = \log \Bigg(\frac{|D|}{\textrm{DF}}\Bigg) + 1,
\end{equation}
where $D$ is the total number of documents and $DF$ is the number of documents in which the term $t$ appears. Finally, the tf-idf weight is computed as
\begin{equation}
\label{eq:tfidf}
\textrm{tf-idf}(t,d,D) = \textrm{tf}(t,d) \cdot \textrm{idf}(t,D).
\end{equation}
In order to get the representative vector of each sentence, we calculate the tf-idf value for each of its content words.

\subsection{Network creation}

\begin{figure}[ht]
 \centering
 \includegraphics[scale=0.35]{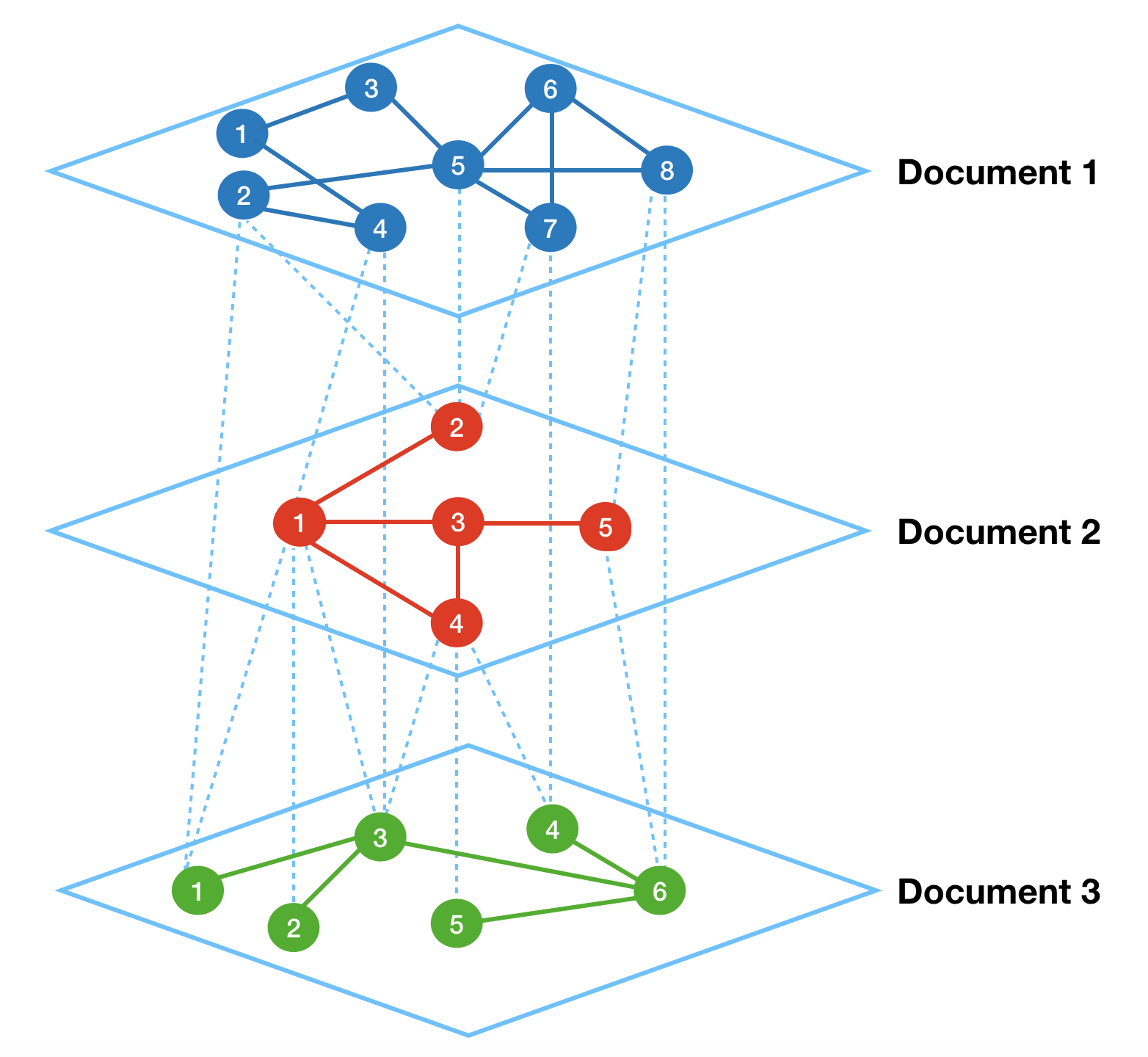}
 \caption{Network model  adopted in this work. Each layer represents a document. Continuous lines are edges linking sentences from the same document (intra-layer), while dashed lines link sentences from different documents (inter-layer). }
 \label{fig:mln_example}
\end{figure}

The multilayered representation of documents is created using the following steps:

\begin{enumerate}

\item \textit{Tf-idf based network creation}: in this model, we first calculate the tf-idf vector representations for each document sentences. Next, each node is represented by a sentence and edges between two sentences $i$ and $j$ are established based on the cosine similarity ($w_{i,j}$) between the corresponding tf-idf vectors. Figure~\ref{fig:tf_idf_net} shows an example of the tf-idf network generated from the example in Table~\ref{tab:procesamiento}. 
\begin{figure}[ht]
 \centering
 \includegraphics[scale=0.32]{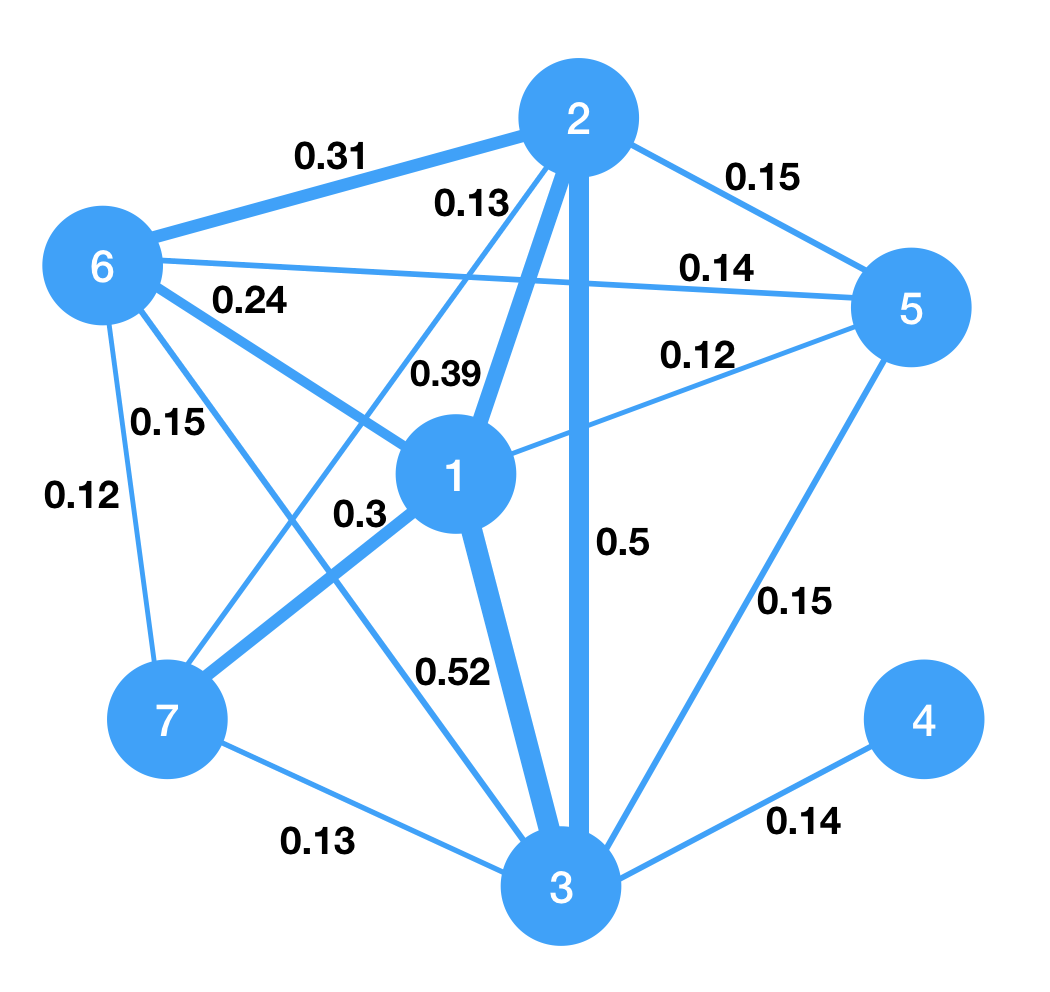}
 \caption{Example of tf-idf based network. Each node number represents the sentences from the piece of text shown in Table~\ref{tab:procesamiento}. Edge weights are created according to the cosine similarity between sentences. 
 }
 \label{fig:tf_idf_net}
\end{figure}

\item \textit{Edge type identification}: we defined two edge types: (i) edges connecting sentences from the same document (intra-layer edges) and (ii) edges connecting sentences from different documents (inter-layer edges). This differentiation is essential to assign relevance to the sentences according to the types of links established.

\item \textit{Type-based edge weighting}: in multi-document summarization, it is important to consider multi-document relationships~\cite{mln}\cite{mln2}. Here, we emphasize the importance of multi-document interrelationships by reinforcing inter-layer edges. Such a reinforcement is done using the a simple linear function:
\begin{equation} \label{eq.alpha}
	\tilde{w}^{(\textrm{inter})}_{i,j} = \alpha  {w}^{(\textrm{inter})}_{i,j}
\end{equation}
where $\alpha$ is a factor that reinforces inter-layer connections (if $\alpha>1$) and
$w_{i,j}^{(\textrm{inter})}$ is the original inter-layer edge weight connecting nodes $i$ and $j$. As we shall show, reinforcing intra-layer links may also be important to improve the characterization of the set of documents. Such an effect is simulated by considering $\alpha<1$ in the experiments. 

\item \textit{Edge removal for non-weighted measurements}:
this step is required because some network measurements are not defined for weighted networks. Here, we removed a fraction $r$ of the weakest links. The remaining edges are considered as unweighted.

\end{enumerate}
To illustrate the creation of multilayer networks, Figure \ref{fig:mln_example} shows an example of a multilayer network generated from a cluster of three documents.

\subsection{Network measurements}

In the summarization context, the goal of a centrality network measurement is to rank the nodes according to its relevance.  The importance assigned by network measurements allows us to determine which are the best ranked sentences that could compose the final summary. Therefore, in this stage, we use a set of network measurements to rank each node network.  Every measurement is used individually, thus, each metric generates one summary.
In this work, we used not only traditional network measurements such as degree, strength, shortest paths, and PageRank; but also  additional measurements to take into account both the topological structure of the networks and their dynamical behavior.
The latter can be captured by considering dynamical processes occurring on the top of the networks. A well-known dynamics used in the context of text analysis in the traditional random walk~\cite{Ramage:2009:RWT:1708124.1708131}.
Here we also use the self-avoiding random walk~\cite{MASUDA2017}, an exploratory dynamics used to define measurements such as accessibility and symmetry. Below we detail each of the measures used in this work.

\begin{itemize}

  \item Degree ($k$): The degree of a node $i$ is the number of edges
  linked to that node. A high degree value suggests that a sentence is related to several others in the document or collection.

\item Strength ($s$): for weighted networks, the strength of a
  node $i$ is the sum of the weights of all its incident edges. In unweighted networks, the strength corresponds to the node degree.

\item Shortest paths ($l$): A shortest path between two nodes $i$ and $j$ is one of the paths that connects these nodes with a minimum length. The length of such a path is henceforth denoted as $d_{ij}$. The average shortest path length is defined as $l_i = \sum_{j\neq i} d_{ij}/ (N-1)$, where $N$ is the total number of words in the network. A sentence is considered relevant for a document or collection if its average distance to any other sentence takes low values. In textual networks, the shortest path length has been useful to identify relevant textual concepts~\cite{1367-2630-13-12-123024}.


\item PageRank ($\pi$): this measurement considers that a node $i$ is relevant if it is connected to other relevant nodes. It can be computed in a recursive way as
\begin{equation} \label{eq:pr1}
  \pi_i = \gamma \sum_j a_{ij} \frac{\pi_j}{k_j} + \beta,
\end{equation}
where $\gamma$ and $\beta$ are damping factors taking values between 0 and 1. In text analysis, this measurement has been used to identify the most probable sense of ambiguous constructions~\cite{Agirre:2009:PPW:1609067.1609070}. A similar measurement based on the number of shortest paths has also been used to gauge similarity in texts modeled as complex networks~\cite{AMANCIO20124406}.

\item Accessibility ($a$): the accessibility quantifies how many nodes are accessible from an initial node when a self-avoiding random walk of length $h$ is performed~\cite{accesibilidad}. This measurement has been
employed to identify borders in geographical networks and to characterize the interplay between structure and dynamics in a wide range of complex networks. Differently from other traditional graph measurements, the accessibility takes into consideration different levels of hierarchy, which can be set by specifying the length $h$ of the random walks.
In textual analysis, this measurement has proven useful to identify key concepts in stylometric applications~\cite{10.1371/journal.pone.0136076}.
To calculate the accessibility, let $p^{(h)}(i,j)$ represent the probability of reaching a node $j$ from $i$ through a self-avoiding random walk of length $h$~\cite{accesibilidad}. The accessibility is defined as the exponential of the true diversity of $p^{(h)}(i,j)$:
\begin{equation}
	a_{i}^{(h)} = \exp \Bigg(-\sum_{j} p^{(h)}(i,j) \log p^{(h)}(i,j)\Bigg).
\end{equation}
It can be shown that the accessibility can be interpreted as the effective number of accessed nodes in the considered dynamics. To illustrate the accessibility concept, we show in Figure \ref{fig:acc} a subgraph created around node $A$. Two configurations of links are considered: (i) continuous red links; and (ii) scenario (i) +  two additional blue dotted links. Note that in the first scenario, all nodes in the second hierarchical level are reached with the same probability ($p=1/5$). This leads to the maximum accessibility value $a_A^{(h=2)} = 5$. In the second scenario, the access to the nodes becomes uneven because two additional paths are created (see dashed blue lines). Such an irregular distribution leads to a decreased value of accessibility $a_A^{(h=2)} = 4.71$.

\begin{figure}[ht]
 \centering
 \includegraphics[scale=0.65]{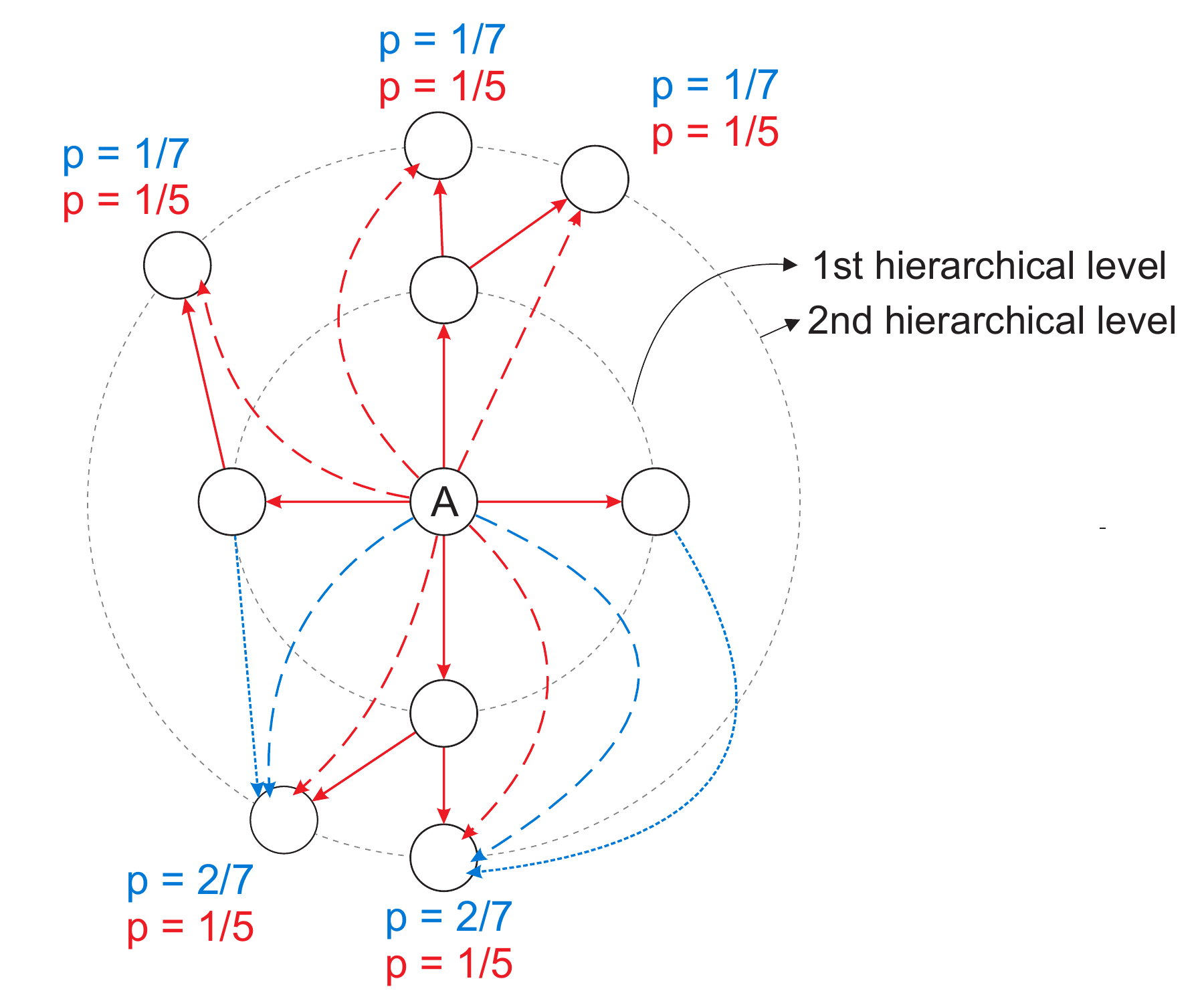}
 \caption{Example of computation of accessibility considering two configuration of edges: (i) red continuous edges only and (ii) red continuous and blue dotted  edges. In (i), the access to the nodes at the second hierarchical level is uniform. Therefore, the accessibility reaches its maximum value $a_A^{(h=2)} = 5$. In (ii), the access to the nodes becomes uneven, as some nodes tend to be accessed more frequently than others. As a consequence, the accessibility of A drops to $a_A^{(h=2)} = 4.71$. }
 \label{fig:acc}
\end{figure}


\item Generalized accessibility ($\tilde{a}$): The accessibility measurement depends on the parameter $h$ for its calculation. The new version of accessibility, called generalized accessibility, is an improvement of accessibility because this new version can be computed without any prior choice of length $h$. Actually, this metric is computed by considering \emph{all lenghts} in the random walk dynamics.
The generalized accessibility depends upon the stochastic matrix $\mathbf{P}$, whose element $p(i,j)$ represents the probability of a random walker to go from node $i$ to $j$ in the next step of the random walk. The transition probability considering all lengths can be computed as
\begin{equation}
	\mathbf{P}^{(\infty)} = \frac{1}{e} \sum_{j=0}^\infty \frac{1}{j!} \mathbf{P}^j.
\end{equation}
The transition probabilities obtained in $\mathbf{P}^{(\infty)}$ can then be used to compute the generalized accessibility using the true diversity of $p^{(\infty)}$, i.e.
\begin{equation} \label{eq.acccc}
	\tilde{a}_i = \exp \Bigg(-\sum_{j} p^{(\infty)}(i,j) \log p^{(\infty)}(i,j)\Bigg),
\end{equation}
where $p^{(\infty)}$ is an element of $\mathbf{P}^{(\infty)}$. This metric has been employed with success as a centrality measurement applied in other text classification tasks~\cite{simetria}.

\item Symmetry ($S$): the network symmetry is a normalized version of accessibility, where the number of accessible nodes is used as normalization factor~\cite{simetria}. The symmetry uses the concept of \emph{concentric level} ($h$) of a node $i$ (see Figure \ref{fig:acc}), which is defined as the set of nodes $h$ hops away from $i$.
Because the main objective of this metric is to quantify how diverse is the
exploration of a neighborhood, the symmetry measurement considers that at each step, the random walker access the next concentric level. Thus, to compute probabilities transitions, all links connecting nodes in the same concentric level are disregarded.
The symmetry is calculated as:
\begin{equation}
  \label{eq:sym}
  S_i^{(h)} = \Big{|} \xi_i^{(h)} \Big{|}^{-1} \exp \Bigg{(}-\sum p^{(h)}(i,j) \log p^{(h)}(i,j) \Bigg{)},
  \end{equation}
where $\xi_i^{(h)}$ is the set of accessible nodes that are at a distance $h$ from the node $i$, i.e. the number of nodes in the $h$-th level. Because this measurement has never been used for summarization, in the current work we evaluated the performance of selecting the sentences with the \emph{highest} and \emph{lowest} values of symmetry.

\item Absorption time ($\tau$): the absorption time is defined as the time it takes for a particle in an internal node to reach an output (absorbent) node through a random walk. This metric quantifies how fast a randomly-walking particle is absorbed by output vertices, assuming that the particle starts the random walk at the input node~\cite{absortion}. The  stochastic transition matrix $\mathbf{P}$ is used to compute this metric. The absorption time is defined using the matrix $\Psi = (\mathbf{I} - \Theta )^{-1}$, where $\mathbf{I}$ is the identity matrix and $\Theta$ is a submatrix of $\mathbf{P}$ which represents the transitions between transient nodes (i.e. non absorbent nodes). It can be shown that the time spent in transient nodes can be computed as
\begin{equation}
	\label{eq:at4}
	t_{i} = \sum_{j} \Psi(i,j).
\end{equation}
For each pair $(i,j)$ of nodes, we define $i$ as a starting node and $j$ as an absorbent node to compute $\Psi$. Thus, here the absorption time of a node $i$ is computed as
\begin{equation}
	\label{eq:at4}
	\tau_{i} = \langle t_i \rangle = \frac{1}{n-1} \sum_{k\neq i} t_k.
\end{equation}
By using this measurement, we expect that sentences taking  low values of absorption time are more semantically related to the other sentences in the text. Thus, to generate the summary, we select the sentences with the lowest values of $\tau_i$.

\end{itemize}

%

\subsection{Summary generation} \label{sum-sel}

In this stage, the final summary is created by selecting the best ranked sentences according to some measurement and strategy selection. The selection can be made by choosing either the highest or the lowest values of the considered metrics. The strategy used for each measurement is summarized in Table \ref{tab:medidas}. For summary generation, a compression rate must be specified. The size of the generated summaries in the current work is the same as the size of the available reference summaries for the corpora used.
\begin{table}[ht]
\centering
\caption{Adopted network measurements for MDS. The weighted version of the networks was considered only with the most traditional measurements.}
\label{tab:medidas}
\begin{tabular}{|c|c|c|}
\hline
\textbf{Selection strategy}& \textbf{Measurement} & \textbf{Abbr.} \\ \hline
\multirow{6}{*}{Highest values} & Degree & dg  \\
& Strength & stg  \\
& PageRank & pr/pr\_w  \\
& Accessibility & access \\
& Symmetry & sym \\
& Gen. Accessibility & gAccess \\ \hline
\multirow{3}{*}{Lowest values} & Shortest Paths & sp/sp\_w  \\
 & Symmetry & sym \\
 & Absorption Time & absT \\ \hline
\end{tabular}
\end{table}

An additional important issue in summarization is the so-called \emph{redundancy treatment}. In the context of automatic summarization, redundancy arises when identical or similar sentences composes the final summary. In network terms, two sentences in the final summary are considered redundant if they are connected by strong links~\cite{Ribaldo2012}. In this paper, we adopted two anti-redundancy detection methods~\cite{Ribaldo2012,ngrams}. In both methods, a similarity threshold is established to compare sentences.
At each step, if the current best ranked sentence is similar to any of the previously selected sentences in the summary, then it is considered redundant. Therefore,  the redundant piece of text is not used to compose the final summary. In this case, the summarization process resumes and the next candidate sentence is evaluated. In the first anti-redundancy method (AR1), the threshold value ($L_1$) is computed as
\begin{equation}
	L_1 = \frac{\max(\sigma(i,j))-\min(\sigma(i,j))}{2},
\end{equation}
%
where $\sigma(i,j)$ is the cosine similarity between two sentences $i$ and $j$.
The second anti-redundancy method (AR2) considers the following similarity measurement:
\begin{equation} \label{ar2eq}
	\tilde{\sigma}(i,j,n) = \sum_{k=1}^n \gamma_k \cdot \frac{|g(i,k) \cap g(j,k)|}{|g(i,k) \cup g(j,k)|},
\end{equation}
where $n$ is the number of $n$-grams to be considered, $g(i,k)$ is the set of $k$-grams of sentence $i$, and $\gamma_k$ is the weight associated with the $k$-gram similarity of two sets. We chose as threshold for $\tilde{\sigma}$ the value $L_2 = 0.1$~\cite{ngrams}. The other parameters in equation \ref{ar2eq} were chosen in accordance with ref.~\cite{ngrams}.

\section{Results}\label{section:results}

In this section, we evaluate the performance of the multilayered approach considering different weighting schemes for inter-layer edges. We also assess the relevance of each network metric for the adopted network representation. Finally, we analyze the correlation between the network metrics in order to identify possible correspondences (equivalences) between the adopted multilayer network measurements.

\subsection{Performance analysis}

The systems were evaluated using the following corpora: CSTNews, a corpus of Portuguese journalistic texts, and DUC (2002 and 2004), which are corpora comprising English journalistic texts. The evaluation of the informativeness of our method was done by using the automatic evaluation method ROUGE-1 \cite{Rouge}, which compares automatically generated summaries and reference texts. This metric was used here because it has been claimed that there is a strong correlation between the ROUGE index and manual (human) judgment~\cite{Ribaldo2012}.

An important parameter in the analyzed multilayer networks is the setup of relevance weights for inter-layer edges (see $\alpha$ in equation~\ref{eq.alpha}). The parameter $\alpha$ accounts for assigning a higher ($\alpha>1$) or lower ($\alpha<1$) relevance for inter-document relationships. The values of $\alpha$ varied in the range $0.5 < \alpha < 1.9$. For each value of $\alpha$, we removed a fixed amount $r$ of the weakest links ($r= \{0.1, 0.2, 0.3, 0.4,0.5\}$). Concerning the anti-redundancy methods, both strategies (AR1 and AR2) considered in this article displayed similar performance (result not shown). For this reason, hereafter we only report the best results.

\begin{figure*}[t]
 \centering
 \includegraphics[scale=0.43]{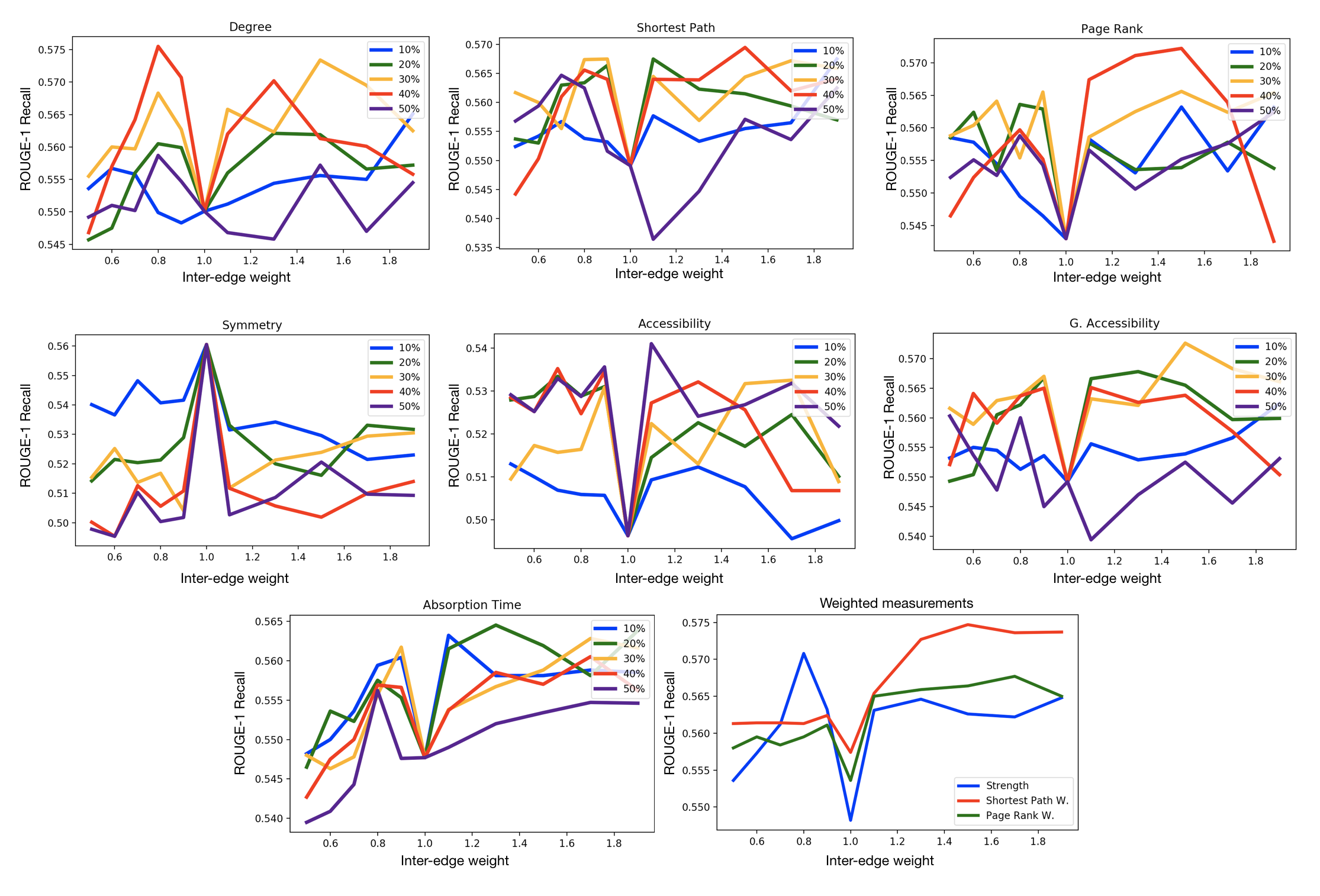}
 \caption{Performance analysis (ROUGE-1 recall) of the adopted multilayer approach in the CSTNews corpus for Portuguese MDS. Each subfigure shows the performance of each proposed network measurement as a function of the inter-layer edge weight parameter ($\alpha$). }
 \label{fig:comparatives_ptg}
\end{figure*}

Figure \ref{fig:comparatives_ptg} shows the overall performance of the multilayer approach (as a function of $\alpha$)  for the CSTNews corpus. For each subplot, we show the curves for different percentages $r$ of removed edges. Remarkably, apart from the simmetry measurement, we observed an improvement in performance when $\alpha \neq 1$. We note that the best results can be obtained in two distinct scenarios: (i) $\alpha < 1$, where a higher relevance is given for intra-document relationships; and (ii)  $\alpha > 1$, where a higher relevance is assigned for the inter-document relationships. The second scenario seems to be more important for improving the performance of the system, since (i) holds just for the degree. We also note from the figures that the best value of $r$ depends on the measurement being analyzed.

\begin{figure*}[t]
  \centering
  \includegraphics[scale=0.49]{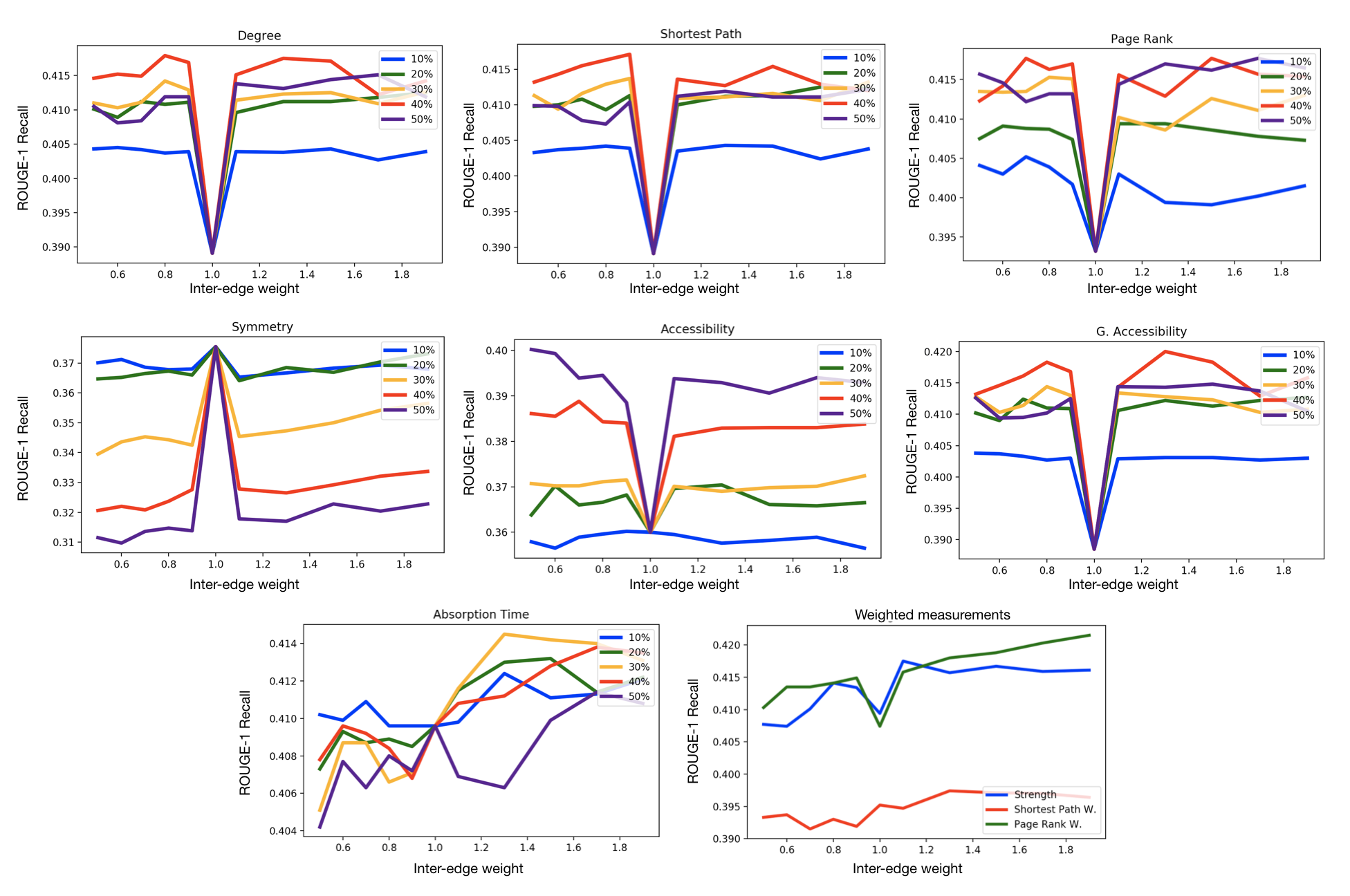}
  \caption{Performance analysis of DUC-2002 corpus for English MDS. Each sub figure shows the performance of each proposed network measurement considering the inter-layer edge weight parameter and thresholds of edge removal. x-axis represents the inter-layer edge weight parameter ($\alpha$) and y-axis is the average ROUGE-1 Recall (RG-1).}
  \label{fig:comparatives_eng}
 \end{figure*}

Figure \ref{fig:comparatives_eng} shows the  performance of the multilayer approach for the DUC-2002 corpus. As observed in the CSTNews corpus (Figure \ref{fig:comparatives_ptg}), a value of $\alpha \neq 1$ is able to improve the performance of the systems, except for the symmetry measurement. Differently from the previous analysis, here the intra-document relationships seems to play an important role for a larger number of network measurements. Optimized results were obtained for $\alpha < 1$ for the degree, shortest path length, PageRank and accessibility (both $a$ and $\tilde{a}$). A major improvement in performance for $\alpha > 1$ was observed for the absorption time measurement. Similarly to the behavior observed for the Portuguese corpus, here the best value of $r$ depends on the adopted network measurement. A similar result was obtained with the DUC-2004 corpus ({see Figure S1 of the Supplementary Information (SI)}), with the best results being obtained for $\alpha \neq 1$ (except for the symmetry).
%

Tables \ref{tab:results_ptg} and \ref{tab:results_eng} summarize the results obtained in both CSTNews and DUC-2002. For the Portuguese case, the best results were obtained with degree and shortest paths, both considering the AR2 strategy to address the anti-redundancy problem. Actually, apart from the generalized accessibility, the best results were always obtained with the AR2 technique. For the English corpus DUC-2002, the PageRank strategy outperformed the other network measurements. A good performance was also observed when the generalized accessibility was used. Here, the AR1 technique for anti-redundancy treatment achieved the best results. Considering the measurements based on self-avoiding random walks, the best performance occurred for the generalized accessibility. This measurement outperformed its hierarchical version probably because the adopted generalization grasps more  information about the network organization. Note that the definition of the generalized accessibility, differently from the hierarchical version, considers \emph{all hierarchical levels} to estimate the effective number of accessed nodes (see equation~\ref{eq.acccc}). Consistently, the symmetry measurement displayed low performance in all adopted datasets. This might be related to the fact that such measurement mainly quantifies the \emph{diversity} of links weights, and not the \emph{proeminence} of nodes. While such a \emph{diversity} might be of interest to quantify particular textual phenomena (see e.g.~\cite{AMANCIO20124406} for an application on authorship recognition), the obtained results suggests that the most important information to quantify relevance is not captured by the symmetry. The results obtained for DUC-2004 (see Table S1) also confirms that low values of $\alpha$ optimizes the performance for particular several measurements.



\begin{table}[ht]
\centering
\caption{Best results obtained for Portuguese MDS.  The network measurements are ranked according to the obtained ROUGE-1 Recall (RG-1).  For each metric, we show the respective parameters for inter-layer edge weight ($\alpha$) and the threshold $r$ for edge remotion.
The anti-redundancy detection (ARD) method that obtained the best result is also shown.}
\label{tab:results_ptg}
\begin{tabular}{|c|l|cccc|}
\hline
   & \textbf{Meas.} & \textbf{$\alpha$} & \textbf{$r$} & \textbf{ARD} & \textbf{RG-1} \\ \hline
1  & dg & 0.80 & 0.40 & AR2 & 0.5755 \\
2  & sp\_w & 1.50 & -- & AR2 & 0.5747 \\
3  & gAccess & 1.50 & 0.30 & AR1 & 0.5726 \\
4  & pr & 1.50 & 0.40 & AR2 & 0.5722 \\
5  & stg & 0.80 & -- & AR2 & 0.5708 \\
6  & pr\_w & 1.70 & -- & AR2 & 0.5677 \\
7  & sp & 1.90 & 0.10 & AR2 & 0.5675 \\
8  & absT & 1.30 & 0.20 & AR2 & 0.5645 \\
9  & sym & 1.00 & -- & AR2 & 0.5605 \\
10 & access & 0.90 & 0.50 & AR2  & 0.5356 \\ \hline
\end{tabular}
\end{table}

\begin{table}[ht]
\centering
\caption{Best results for English MDS (DUC-2002 Corpus). The network measurements are ranked according to the obtained ROUGE-1 Recall (RG-1).For each metric, we show the respective parameters for inter-layer edge weight ($\alpha$) and the threshold $r$ for edge remotion.
The anti-redundancy detection (ARD) method that obtained the best result is also shown.}
\label{tab:results_eng}
\begin{tabular}{|c|l|cccc|}
\hline
   & \textbf{Meas.} & \textbf{$\alpha$} & $r$ & \textbf{ARD} & \textbf{RG-1} \\ \hline
1 & pr\_w & 1.90 & -- & AR1 & 0.4215   \\
2 & gAccess & 1.30 & 0.40 & AR1 & 0.4200  \\
3 & dg & 0.80 & 0.40 & AR1 & 0.4179   \\
4 & pr & 1.70 & 0.50 & AR2 & 0.4177   \\
5 & stg & 1.10 & -- & AR1 & 0.4175 \\
6 & sp & 0.90 & 0.40 & AR1 & 0.4171   \\
7 & absT & 1.30 & 0.30 & AR1 & 0.4145  \\
8 & access & 0.50 & 0.50  & AR2 & 0.4002   \\
9 & sp\_w & 1.30 & -- & AR2 & 0.3974   \\
10 & sym & 1.00 & -- & AR2 & 0.3755   \\ \hline
\end{tabular}
\end{table}

The results observed in the above experiments suggests that the multilayer representation is relevant to identify the most prominent nodes (sentences) in documents. The importance of the intra- and inter-layer edges seems to be dependent on the  dataset, language and metric used to rank sentences. This was clear when we observed that both scenarios ($\alpha>1$ and $\alpha<1$) are possible, even when considering the same language. While such a multilayer representation has been implicitly used in some textual contexts, no discrimination between intra- and inter-layer edges has been considered for summary generation~\cite{Ribaldo2012}. Given the effectiveness of the adopted representation to optimize the performance obtained by network measurements, we believe that it could be considered in other related applications where intra- and inter-relationships are relevant. This is the case, e.g. of applications related to text mining and identification of key concepts in scientific areas~\cite{SILVA2016487}.

This study focused on the \emph{evaluation} of multilayer-based network approaches for MDS. Even though we aimed at studying the relevance of discriminating intra- and inter-layer edges in multilayer networks, it is still interesting to compare the efficiency of statistical approaches (such as the one evaluated in the current study) and other approaches dependent upon more informed linguistic data. For comparison purposes, we show in Table \ref{tab:other_results} a set of other works that achieved the best results for the corpus we used. For Portuguese MDS the works are: GistSumm~\cite{Ribaldo2012}; BushyPath and Depth-first Path systems~\cite{Ribaldo2012}; and CSTSumm, which follows a strategy based on cross-document structure theory~\cite{cst2}. In the case of English MDS, we considered the following works: DUC-best, which is the system with highest ROUGE scores for DUC conferences; BSTM~\cite{wang2009multi}, which uses a  Bayesian sentence-based topic model for summarization; FGB~\cite{wang2008integrating}, which proposes  a new language model to simultaneously cluster and summarize the documents; and LexPR~\cite{erkan-radev:2004:EMNLP}, which constructs a sentence connectivity graph based on cosine similarity and selects important sentences based on the eigenvector centrality. Considering the Portuguese language, the multilayer approach outperformed several other statistical methods. The best result was obtained, however, with the GistSumm technique, which considers more linguistic and semantical information than the multilayer approach. For both English corpora, linguistic dependent approaches outperformed the method based on multilayers. However, in the DUC-2004 dataset, only a minor difference between the multilayer and other methods was observed.

\begin{table}[ht]
\centering
\caption{List of works for Portuguese and English MDS with the respective average ROUGE-1 Recall (RG-1) scores. The best results of the multilayer approach evaluated in this work are highlighted.}
\label{tab:other_results}
\begin{tabular}{|lc|lc|lc|}
\hline
\multicolumn{2}{|c|}{\textbf{CSTNews}} & \multicolumn{2}{c|}{\textbf{DUC-2002}} & \multicolumn{2}{c|}{\textbf{DUC-2004}} \\ \hline
\textbf{System} & \textbf{RG-1} & \textbf{System} & \textbf{RG-1} & \textbf{System} & \textbf{RG-1}  \\ \hline
GistSumm & 0.6643 & DUC-best & 0.4986 & BSTM & 0.39065 \\
\textbf{Multilayer} & \textbf{0.5755} & BSTM & 0.48812 & FGB & 0.38724 \\
BushyPath & 0.5397 & FGB & 0.48507 & DUC-best & 0.38224 \\
Depth-first Path & 0.5340 & LexPR & 0.47963 & LexPR & 0.3784 \\
CSTSumm & 0.5065 & \textbf{Multilayer} & \textbf{0.4215}  & \textbf{Multilayer} & \textbf{0.3707} \\ \hline
\end{tabular}
\end{table}

\subsection{Correlation analysis}

\begin{figure}[ht]
 \centering
 \includegraphics[scale=0.48]{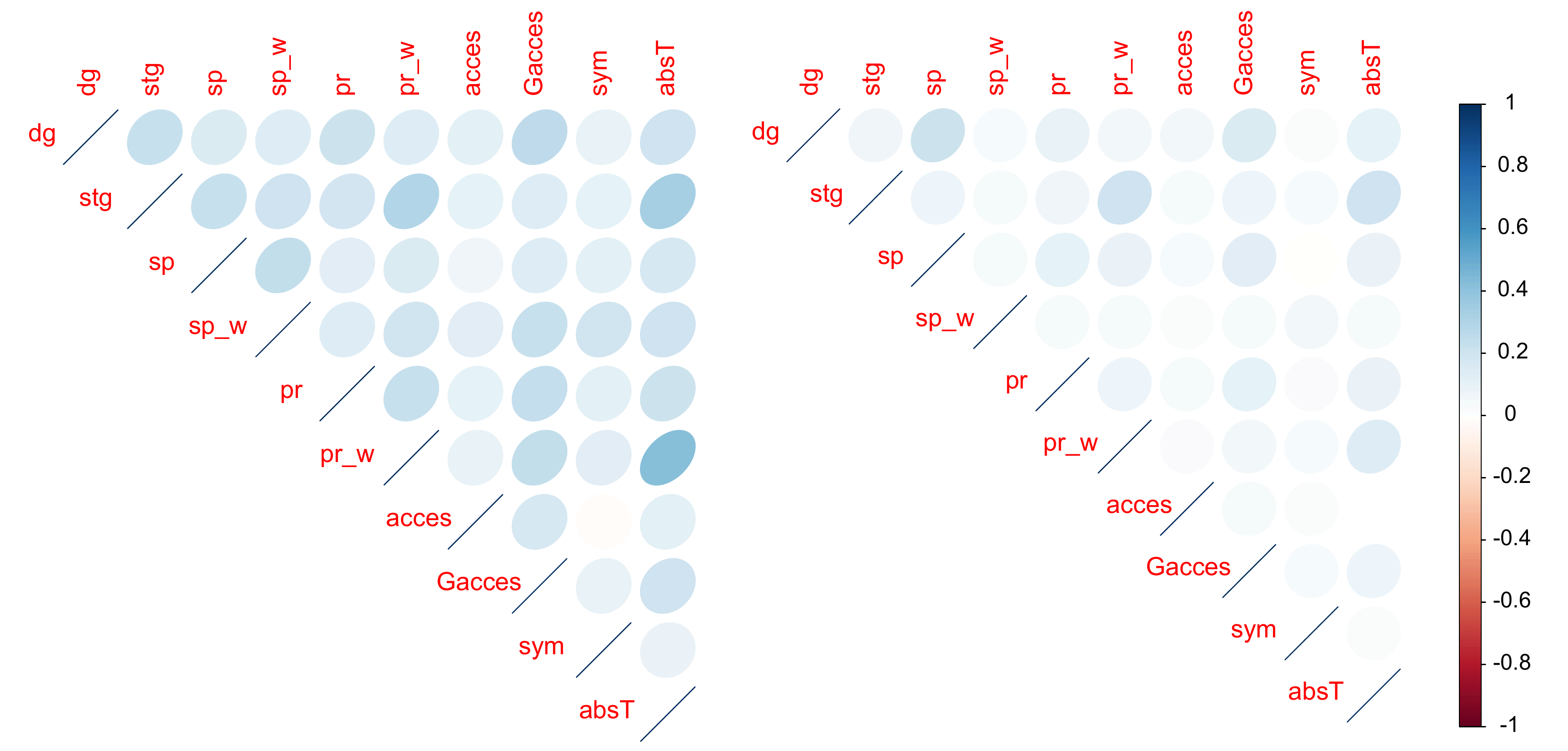}
 \caption{Spearman rank correlation coefficients for the considered network measurements, considering both CSTNews corpus (left) and DUC-2002 corpus. Note that, in general, there is a weak correlation between the adopted measurements. The strongest correlation occurs for the pair weighted PageRank and absorption time, for the Portuguese corpus.}
 \label{fig:correlations}
\end{figure}

In Figure \ref{fig:correlations}, we show the Spearman rank correlation coefficient observed for both CSTNews (left) and DUC-2002 (right) corpora. For the CSTNews case, in general, the correlation are not strong, which means that the metrics are not equivalent in the summarization task. The highest correlation, however, occurs for the pair weighted PageRank and absorption time. Such a similarity might occur because both measurements are based on the behavior of a random-walk dynamics. For the DUC-2002 corpora, the correlations are even lower, confirming thus an absence of correspondence among the adopted network metrics. Low values of Spearman rank correlation coefficient were also observed in the DUC-2004 corpus {(see Figure S2)}. The complimentary role played by the metrics in the adopted multilayer networks suggests that such metrics could be combined (e.g. in a voting system) to improve the performance of extractive summarizers. Most importantly, the differentiation of edge types in characterizing collections of documents should be taken into account whenever semantical links are at the core of the target task or investigation.


\section{Conclusion}\label{section:conclusions}


In this paper, we evaluated the efficiency of a multilayered approach for the extractive summarization problem. Several interesting findings were observed when applying such a simple model to
a set of Portuguese and English texts. We found that the performance observed is increased when intra- and inter-layer edges are considered with distinct relevance. For the Portuguese case, for most of the measurements, the best performance was obtained when inter-document relationships are strenghtened. Conversely, for the English case, both intra- and inter-layer edges yielded the best results. Concerning the metrics, excellent performance was obtained for the degree (Portuguese), weighted PageRank (English) and the generalized acessibility (for both Portuguese and English). Differently from previous results obtained for network-based single document summarization~\cite{lucas}, we found no strong correlation among the adopted measurements.

Given the complimentary role played by the metrics, this study could be extended in order to consider acombination of metrics to improve the performance of the systems. The adopted model could also be extended in future works by using the concept of word embeddings~\cite{embeddings} to enhance the representation of sentences. Finally, future works could combine traditional document summarization techniques and multilayered network representations. Such a combination can be created e.g. by combining network concepts and linguistic features (such as sentence length, number of proper nouns and sentence location) via machine learning or hybrid classifiers~\cite{10.1371/journal.pone.0136076,Leite}.


\section*{Acknowledgments}

J.V.T. acknowledges financial support from CNPq (Brazil). D.R.A. thanks S\~ao Paulo Research Foundation (FAPESP grant no. 16/19069-9) for the financial support.

\bibliography{mybibfile}

\end{document}